\documentclass[12pt]{spieman}  

\usepackage{amsmath}
\usepackage{amsfonts}
\usepackage{amssymb}
\usepackage{graphicx}
\usepackage{setspace}
\usepackage{tocloft}
\usepackage[dvipsnames,table]{xcolor}
\usepackage{tikz}
\usepackage{svg}
\usepackage[super,sort,comma]{natbib}
\usepackage{multirow}
\usetikzlibrary{arrows,positioning,calc}

\tikzset{>=latex}
\def\connect[#1] (#2)!#3!(#4){ 
  \draw[#1] (#2) |- ($(#2)!#3!(#4)$) node[pos=0.5] (#2-#4-1) {}
  -| (#4) node[pos=0.5] (#2-#4-2) {}
}

\title{Evaluating Explainability: A Framework for Systematic Assessment and Reporting of Explainable AI Features}

\author[a,*]{Miguel A. Lago Ph.D.}
\author[a]{Ghada Zamzmi Ph.D.}
\author[a]{Brandon Eich Ph.D.}
\author[a]{Jana G. Delfino Ph.D.}
\affil[a]{U.S. Food and Drug Administration, Silver Spring, MD 20993, USA}

\cftpagenumbersoff{figure}
\cftpagenumbersoff{table} 
\begin{document}
\maketitle








\begin{abstract}

\noindent \textbf{Purpose:} Explainability features are intended to provide insight into the internal mechanisms of an AI device, but there is a lack of evaluation techniques for assessing the quality of provided explanations. We propose a framework to assess and report explainable AI features.

\noindent \textbf{Materials and Methods:} Our evaluation framework for AI explainability is based on four criteria: 1) Consistency quantifies the variability of explanations to similar inputs, 2) Plausibility estimates how close the explanation is to the ground truth, 3) Fidelity assesses the alignment between the explanation and the model internal mechanisms, and 4) Usefulness evaluates the impact on task performance of the explanation. Finally, we developed a scorecard for AI explainability methods that serves as a complete description and evaluation to accompany this type of algorithm.

\noindent \textbf{Results:} We describe these four criteria and give examples on how they can be evaluated. As a case study, we use Ablation CAM and Eigen CAM to illustrate the evaluation of explanation heatmaps on the detection of breast lesions on synthetic mammographies. The first three criteria are evaluated for clinically-relevant scenarios.

\noindent \textbf{Conclusion:} Our proposed framework establishes criteria through which the quality of explanations provided by AI models can be evaluated. We intend for our framework to spark a dialogue regarding the value provided by explainability features and help improve the development and evaluation of AI-based medical devices.

\end{abstract}



\begin{spacing}{1}   


\section{Introduction}
Traditionally, machine learning models relied on hand-crafted features designed by developers for specific tasks. These feature-engineered machine learning systems were easy to understand, since relevant features were engraved into the model architecture. This simplicity made such models comprehensible, but their limited capabilities often left much to be desired. In the last few decades, developments of complex neural networks such as Deep Neural Networks (DNN) and generative artificial intelligence (AI) have led to unprecedented levels of accuracy\cite{guidotti2018survey,he2020extract}. However, the opacity of such models makes it difficult to understand the reasons behind specific predictions or outcomes, creating a ``black box'' effect in which the inner workings of the model are incomprehensible to human users \cite{broniatowski2021psychological-NIST,doshivelez2017rigorousscienceinterpretablemachine,schmidt2019quantifyinginterpretabilitytrustmachine}.

Artificial intelligence (AI) applications deployed in critical domains, such as healthcare, judiciary, and financial markets, raise the need for end users to understand, interpret, and comprehend the logic inside ``black box'' models. Understanding the logic behind AI recommendations is key to their adoption in healthcare settings. A recent workshop held by the American College of Radiology concluded that ``clinical implementation of AI in radiology will continue to be limited until the safety, effectiveness, reliability, and transparency of such tools are more fully addressed''\cite{larson2023acr}. Further, the American Medical Association emphasizes the need for AI tools that focus on transparency and explainability in their designs \footnote{\href{https://www.ama-assn.org/system/files/ama-ai-principles.pdf}{https://www.ama-assn.org/system/files/ama-ai-principles.pdf}}. However, the lack of a precise definition of explainability complicates the implementation of explainable AI methods, as a clear understanding of what explainability entails is important for their effective design and implementation.

Explaining a CNN-based model's reasoning is becoming a very popular topic in the literature.  
\citeauthor{Haar_2023} \cite{Haar_2023} provides a comprehensive review of recent gradient-based, perturbation-based, and approximation-based explainability methods for CNNs.  \citeauthor{ExplainingExplanations}\cite{ExplainingExplanations} conducts a survey of available literature, providing a taxonomy of definitions and focusing on what is being explained by provided model explanations. The authors highlight the absence of standardized assessment methodology for explanations of AI systems, but focus on \textit{development} of explainable AI systems; here our focus is on \textit{evaluation} of explanations provided by AI systems.  

The topic of evaluating explainability has received significantly less interest in the literature. \citeauthor{gunning2021-DARPA-XAI}\cite{gunning2021-DARPA-XAI} concludes that ``\textit{User studies are still the gold standard for assessing explanations}'', which further motivates the need for an evaluation framework that reduces the need for such user studies. A recent review on explainable AI techniques in radiology and nuclear medicine by \citeauthor{vries2023explainable}\cite{vries2023explainable} concludes that ``\textit{There is currently no clear consensus on how explainable AI (i.e., XAI or explainability) should be deployed in order to close the gap between medical professionals and DL (Deep Learning) algorithms for clinical implementation}.'' 

To complicate the field even further, providing an explanation is insufficient; to have value, an explanation must also be useful. To be useful, explanations must balance providing useful information while avoiding overwhelming users with more information than they can process, affecting the trust of end users on the AI device \cite{venkatesh2024gradient}. Showing too much information without enough guidance can create unnecessary cognitive load, confuse users, and degrade human performance. For example, Fuxin et al.\cite{fuxin2021heatmaps} showed that study participants spent an average of 140\% more time to process the visualization and answer the questions when shown SAG (Structured Attention Graphs) explanations, compared with traditional heatmaps which did not change user processing time compared to no visual explanations. Modifying their visualization of SAG to include interactive arrows that highlights images that are similar (but not identical) to the query, decreased participant response times by 30\% while increasing accuracy by 30\%. Similarly, explanations which provide incorrect information can detract from user performance. In a recent deception study with computer aided marks, reader performance suffered (decreased) when readers were presented with intentionally incorrect computer aided marks \cite{bernstein2023can}, underscoring the need to ensure that all information provided to users by AI systems is accurate and beneficial. Other studies have also demonstrated that biased explanations from AI models can lead to decreased diagnostic accuracy \cite{jabbour2023measuring,khera2023automation}. These prior works underscore the need for development of a comprehensive framework for evaluating and reporting the performance of explainability features such as heatmaps, counterfactuals, and confidence scores.

Such an evaluation framework is important for providing insights into the performance of explainability methods across different criteria, which can benefit various user groups. For model developers, a comprehensive evaluation framework can offer a means to analyze strengths and weaknesses of explainability methods, enabling model optimization and bench-marking against industry standards. For model evaluators and regulators, a standardized framework allows for objective comparisons, making assessments more efficient and enhancing credibility by aligning with safety and compliance standards. For end users, such as decision makers and doctors in clinical practice, a comprehensive evaluation framework can help foster trust and confidence in AI-outputs and provide insights to inform clinical choices and support effective patient communication. Providing an explanation of how models arrive at specific recommendations allows doctors to compare the reasoning underlying the AI output against their clinical judgement.

This paper aims to lay the foundation for evaluating explainability features in AI-enabled medical devices. As medical device manufacturers increasingly seek to incorporate explainability features, it is critical to develop robust, well-established evaluation paradigm. As different users have different needs for explinablility features, our approach to defining explainability and its evaluation is specifically framed from the perspective of the model’s end-users: clinicians, radiologists, physicians, etc.rather than model developers.  

\section{Explaining Explainability}

We define \textit{Explainability} as the ability of an AI model to translate its internal reasoning mechanisms in a humanly understandable way. Put another way, explainability refers to a representation of the mechanisms underlying AI systems' operation \cite{ai2023artificial-NIST} that is meant to help understand the model response. Explainable AI can be achieved in two main ways: embedded directly into the architecture of the AI model itself (intrinsic explainability or model interpretability) or as a post-hoc method attached to the model. 

As a user of the model, explainability is welcome but for specific reasons. An explanation of the model's internal mechanisms should be related to the specific task (clinical relevancy) and for a specific expertise field. Users will find that explanations help with the trust they put on the model and the final decision will be based on a combination of the explanation and the model output.

The explanation seeks to describe the process, or rules, that were implemented to achieve a model's output, independent of the context in which that output will be used 
\cite{broniatowski2021psychological-NIST}. Typically, explanations are detailed, technical, and may be causative. Explainable systems can be debugged and monitored more easily, and they lend themselves to more thorough documentation, audit, and governance \cite{ai2023artificial-NIST}. 

Explainability in medical imaging can take many forms \cite{Hossain2023-explainableaiformedicaldata}, including: heatmaps (i.e., saliency maps) that highlight regions of the image that contribute to the model response, prototypes that shows examples of similar inputs, counterfactuals that shows a different or opposite class as an explanation, and confidence ratings that shows a score for the likelihood of an output to be correct. This manuscript will focus on evaluating explainability features of Convolutional-Neural-Network-based models. 

\section{Evaluating Explainability}

In this section, we present a framework for evaluating explainability methods and a reporting mechanism through an explainability card. This card is divided in two sections: 1) a descriptive section containing information about the explainability method; and 2) a quantitative section that assesses the performance of the method using relevant criteria and metrics. The supplementary material provides an example of this card.

Such a card offers several benefits to different stakeholders including end-users and regulators. For example, it can offer clarity and understanding of how different explainability methods perform, enabling informed decision-making, benchmarking, and targeted improvements. It can also ensure quality assurance and facilitate iterative improvement. Additionally, using this card can facilitate improved communication among stakeholders by providing a standardized format for presenting and discussing explainability metrics and methods, ensuring that everyone has access to the same information and can engage in more productive dialogues.

\subsection{Descriptive Information }

The descriptive information covers everything related to the explainability method and the application including an (1) overview of the method, (2) context of use, (3) limitations and recommendations, and (4) validation setting.

The overview provides the name of the method used to generate the explanation, its description and abbreviation, its type (e.g., local, post hoc), citation or reference for further reading, and the software used to implement the method. 

The context of use section identifies the audience, specific task and the model to which the explainability method is applied. Understanding the context of use is important as the effectiveness and relevance of an explainability method can vary depending on the task, and it is hard to fully understand and interpret the meaning of the explainability method without knowing this context. For example, a heatmap is typically used to provide visual explanations for classification models by highlighting important regions in an image that contribute to the model output. However, its application would differ in a regression task, where the focus might be on understanding how different regions of the input contribute to a continuous output variable, and in a segmentation task, where the heatmap would need to explain pixel-level decisions rather than overall image classification.

The limitations and recommendations section provides an overview of the constraints, limitations, and assumptions associated with the explainability method to ensure that stakeholders are well-informed about its practical aspects. There may be cases where the method produces misleading results, particularly in edge cases or under atypical conditions. Therefore, understanding these limitations is necessary for correctly interpreting the outputs of the explainability method. Additionally, it is important to recognize if the explainability method relies on certain underlying assumptions, such as the presence of a stable underlying model, and specific operating conditions. This section also lists specific cases or conditions where the method fails or underperforms. By highlighting these failure cases, users can better anticipate potential issues and take preventive measures. In addition to the limitation(s), recommendations can be included to maximize the utility of the explainability method and integrate the method into existing workflows. 

Finally, the validation setting section should provide details about the design of the validation process. For example, this section can provide information related to user studies. 

\subsection{Quantitative Evaluation}
\begin{figure}
    \centering
    \includegraphics[page=3,trim={7cm 4cm 7cm 4cm},clip,width=1\textwidth]{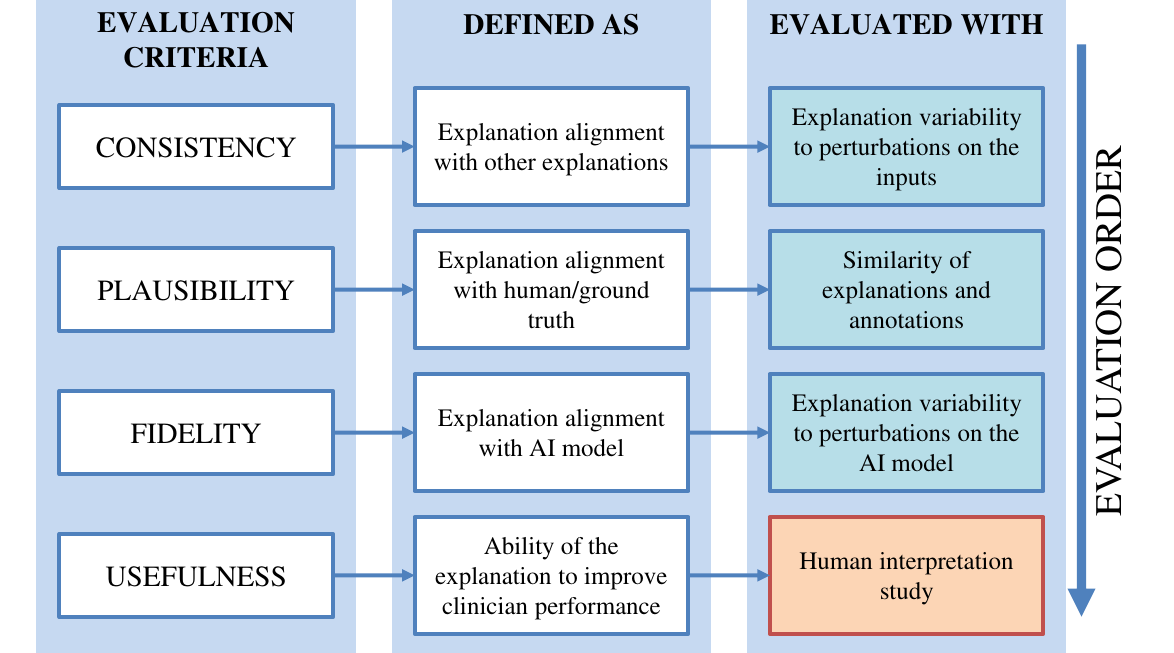}
    \caption{Explainability sources, their definitions, and their evaluation.}
    \label{fig:explainability-definitions}
\end{figure}

To evaluate the performance of explainability methods, we first need to establish a set of evaluation criteria (Fig. \ref{fig:explainability-definitions}). In this section, we define four key evaluation criteria and present different approaches  that can be used to assess their performance. Each criterion is explained in terms of relevance to clinical applications, and we provide demonstrative examples from two explainability methods: Eigen CAM and Ablation CAM.

Eigen CAM \cite{Muhammad2020EigenCAM} and Ablation CAM \cite{Desai2020AblationCAM} are two widely used explainability methods for visualizing how deep learning models make decisions. Eigen CAM highlights important regions by computing the principal components of feature maps, while Ablation CAM systematically removes parts of the model to assess the contribution of each component to the final prediction. These methods are complementary, providing both global and local insights into model explainability. 

We leveraged the M-SYNTH dataset \cite{sizikova2024msynth} which includes synthetic x-ray mammography images. The use of synthetic data allows us to have a known ground-truth and reduce the variability from real images. This allowed us to isolate the analysis of the heatmaps without needing to consider the data and annotations quality in the dataset. More specifically, we used a sample size of 150 images where 50\% contain 1 lesion within them. Using all of the images that contain a lesion allowed us to compare the model's prediction to a ground truth that is perfectly known (i.e., lesion presence and location).

To evaluate the performance of Eigen CAM and Ablation CAM, we applied them to explain the output of a Faster R-CNN model \cite{ren2016faster} trained to detect lesions on digital mammography datasets. The Faster R-CNN was trained on fatty breast density images with a lesion radius of 7 mm. After training, the model had an area under the receiver-operating characteristic (ROC) curve (AUC) value of 1 for the fatty breast density, with decreasing AUC as density increased (the model had the worst performance with the dense density in the M-SYNTH dataset). We used a smaller lesion (5 mm radius) on our testing dataset to reduce the AUC and add variability to the model performance. 

\subsubsection{Consistency}
Consistency measures how reliably an explainability method provides similar explanations when small variations are introduced to the input data. These variations could be noise, slight image rotations, or shifts, which are often seen in real-world applications such as patient movement during imaging. Consistency is also referred to as stability or robustness.

A consistent visual-based explainability method should generate similar explanations even when the input image undergoes small changes. Evaluating the consistency of the explanation provided by explainability methods is important as it builds trust and reliability by providing predictable and dependable explanations, even with minor variations (e.g., noise or patient movement) in input data. Consistency should be the first step toward a good explanation.

To measure consistency, various metrics can be applied to compare the original and transformed images. For example, distance-based metrics can assess the difference between pixel intensities in the heatmaps of the perturbed and non-perturbed images, or compare two feature vectors. Overlap metrics, such as Intersection over Union (IoU), can quantify how much the regions of interest highlighted in the heatmaps overlap between the original and transformed images. Additionally, we can use image similarity metrics such as SSIM (Structural Similarity Index Measure) or MSE (Mean Squared Error). Consistency can be evaluated by comparing explanations for similar cases under different perturbations, such as noise, rotations, or image degradation. These perturbations allow for the quantification of consistency based on how much the explanation changes under varying degrees and types of alterations. A highly consistent explanation would remain relatively unaffected by such perturbations, and would demonstrate stable metric values across different scenarios.

\begin{figure}[t]
    \centering
    \begin{tabular}{ll}
    {\large A) Consistency} & \\
         \includegraphics[trim={0cm 0cm 0cm 3.7cm},clip,width=0.5\linewidth]{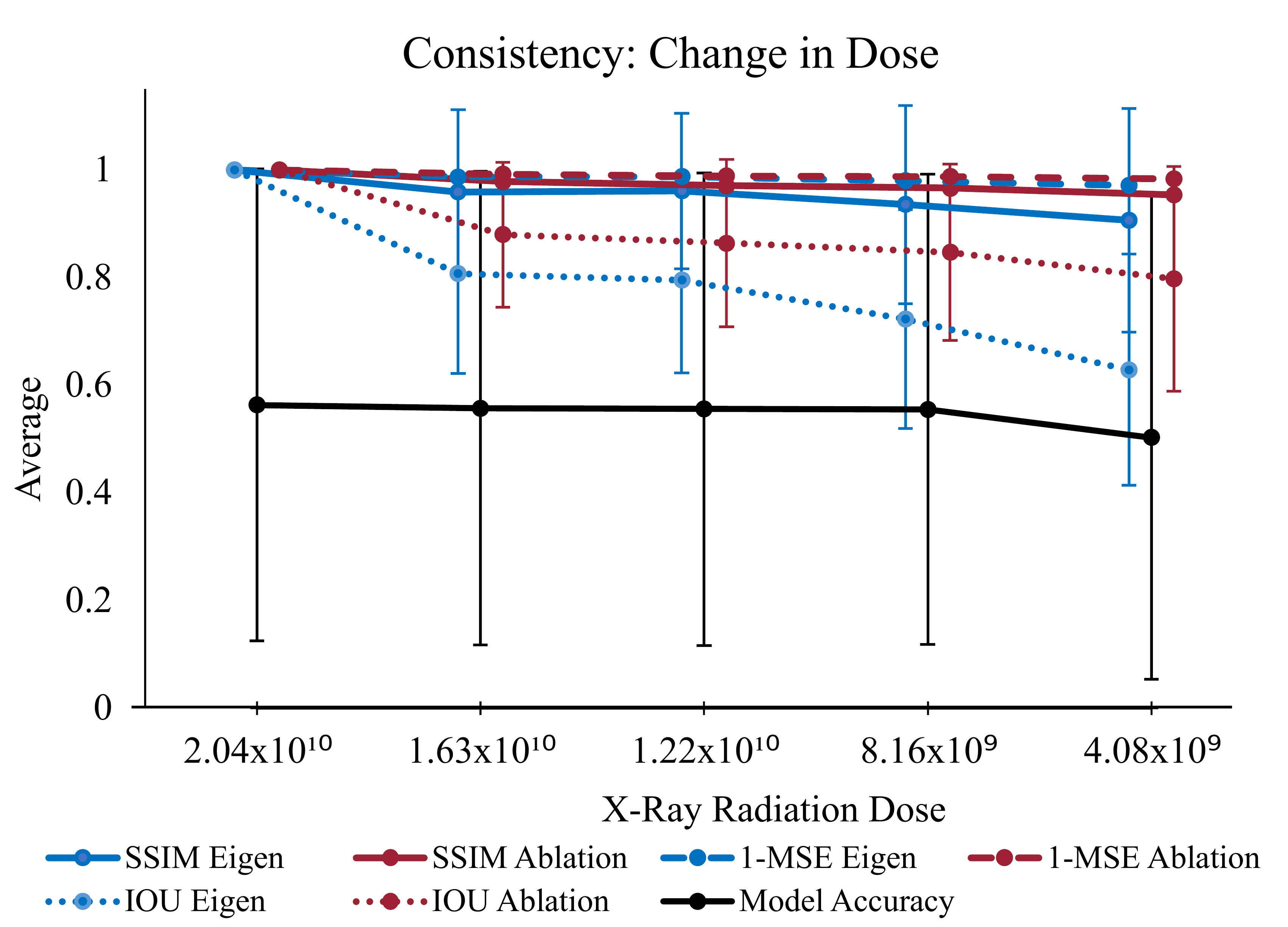}&\includegraphics[trim={0cm 0cm 0cm 3.7cm},clip,width=0.5\linewidth]{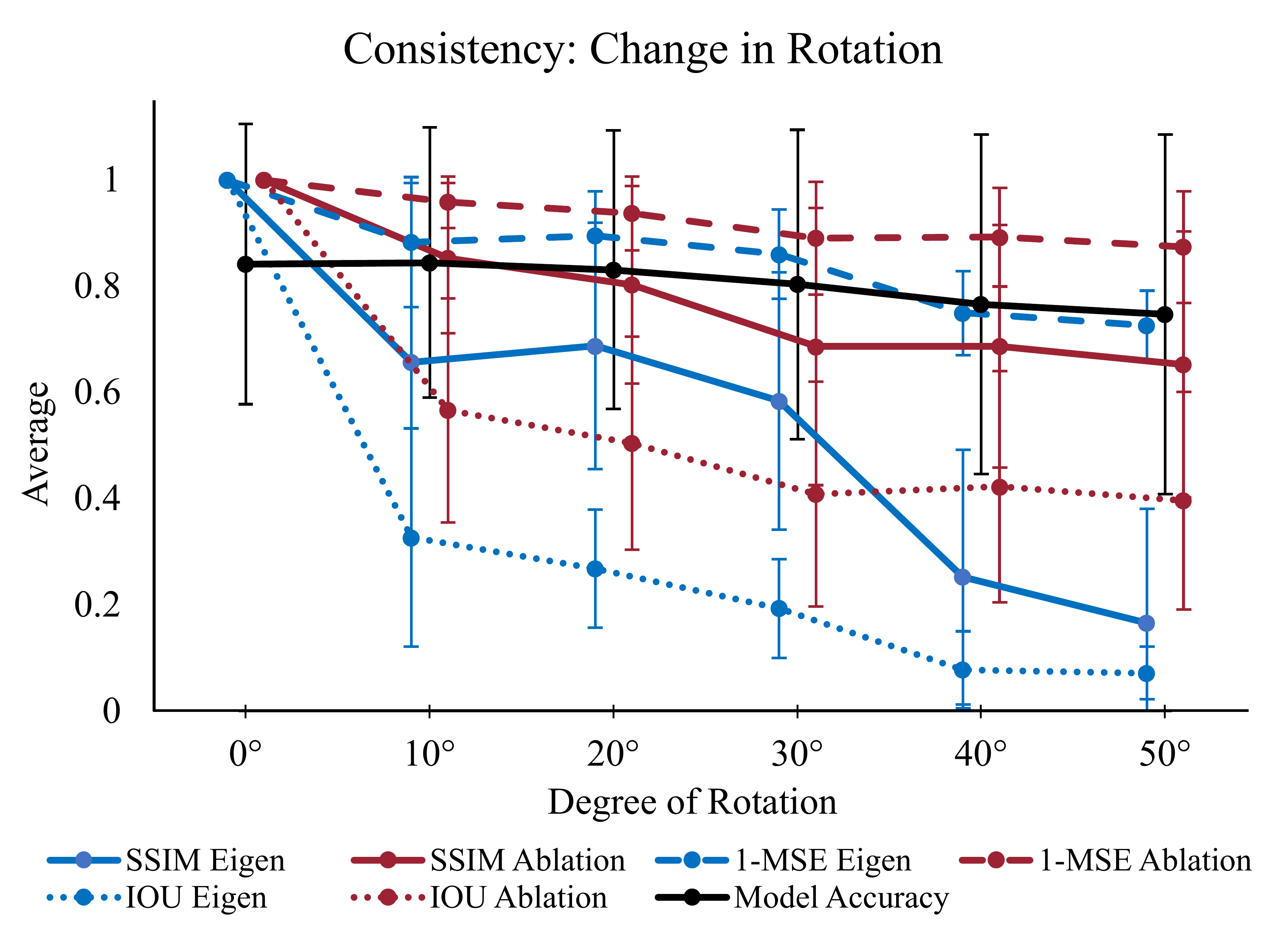}  \\ 
         {\large B) Examples} &\\
         \multicolumn{1}{c}{High Dose} & \multicolumn{1}{c}{Rotation 0$^\circ$} \\
         \multicolumn{2}{c}{\includegraphics[width=0.45\linewidth]{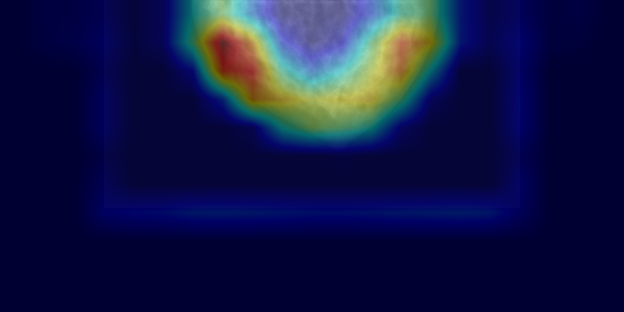}
        \includegraphics[width=0.45\linewidth]{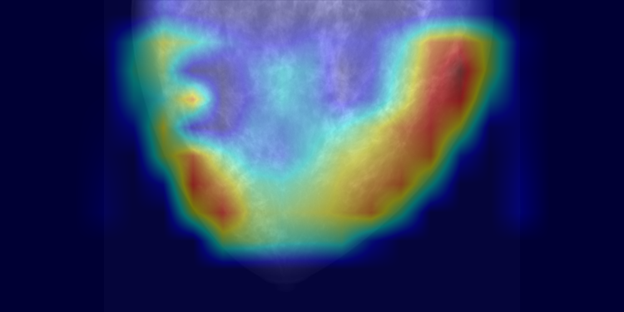}} \\
        \multicolumn{1}{c}{Low Dose} & \multicolumn{1}{c}{Rotation 50$^\circ$} \\
        \multicolumn{2}{c}{\includegraphics[width=0.45\linewidth]{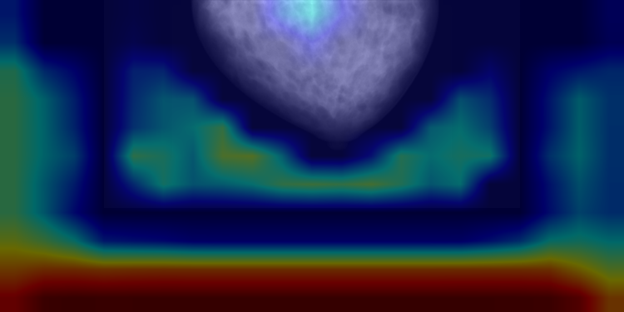}
        \includegraphics[width=0.45\linewidth]{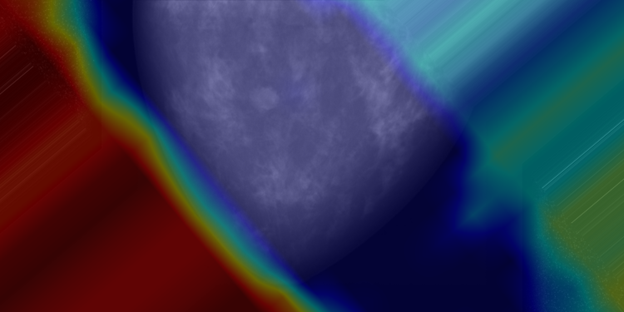}}
    \end{tabular}
    
    \caption{A) Results for Consistency based on Radiation Dose and Rotation. Metrics are calculated using  Structural Similarity Index Measure (SSIM), Mean Squared Error (MSE), and Intersection over Union (IoU). B) Example of a heatmap from two different doses (left) and two different rotations (right) on a synthetic breast mammography.}
    \label{fig:results-main1}
\end{figure}

Figure~\ref{fig:results-main1} illustrates how the consistency of heatmaps, generated using Eigen CAM and Ablation CAM, varies under different conditions in digital mammography. Specifically, the figure explores changes in heatmaps due to variations in X-ray radiation doses and degrees of image rotation. In Figure~\ref{fig:results-main1}.A, consistency is quantified using metrics such as SSIM, MSE, and IoU, with the corresponding values compared across different radiation doses and rotation angles. Despite the model's accuracy remaining stable, SSIM, MSE, and IoU values exhibit notable fluctuations,  which highlights how explainability methods can behave inconsistently under subtle changes. Figure~\ref{fig:results-main1}.B presents visual examples of heatmaps, demonstrating how they change when the radiation dose is reduced and when the image is rotated by 50 degrees. The top left shows the heatmap under a high radiation dose, while the bottom left shows the heatmap at a lower dose. Similarly, the top right displays the heatmap for the original image, and the bottom right shows the heatmap after a 50-degree rotation, where both the input and heatmap were set back to their original position for demonstration purposes. 

This analysis underscores the need to evaluate heatmap consistency under clinically relevant variations, such as changes in radiation dose or image rotation. If small and insignificant variations in the input data can cause large shifts in the explanation, the method becomes difficult to rely on, as it may produce erratic interpretations under real-world conditions. Therefore, consistency should be considered the first step in evaluating any explainability method.

\subsubsection{Plausibility}
Plausibility measures the alignment between the generated explanation and human explanation provided as ground truth labels. For example, in visual-based explanations such as heatmaps, a plausible explanation for detecting pneumonia in a medical image should highlight areas in the lungs where a radiologist would expect to see signs of pneumonia. If the heatmap aligns with areas a radiologist identifies as relevant, it is considered plausible. In non-visual-based explanations, such as feature importance scores in a model predicting patient outcomes based on clinical features, a plausible explanation might highlight high blood pressure and cholesterol as important factors for heart disease, aligning with established medical knowledge or reference standard.

To quantitatively assess the alignment between a model's explanation and domain knowledge (ground truth), various overlap metrics can be employed. For example, metrics like IoU can be used to compare heatmaps or localization outputs with ground-truth object bounding boxes, segmentation masks, or human attention maps in imaging data. Additionally, correlation metrics such as Spearman's rank correlation and rank correlation \cite{vilone2021notions} can be used to evaluate the correspondence between the generated and ground truth explanations.

\begin{figure}[t]
    \centering
    \begin{tabular}{ll}
    {\large A) Plausibility} & {\large B) Examples}\\
         \multirow{2}{*}{\includegraphics[trim={0cm 0cm 3cm 3.5cm},clip,width=0.45\linewidth]{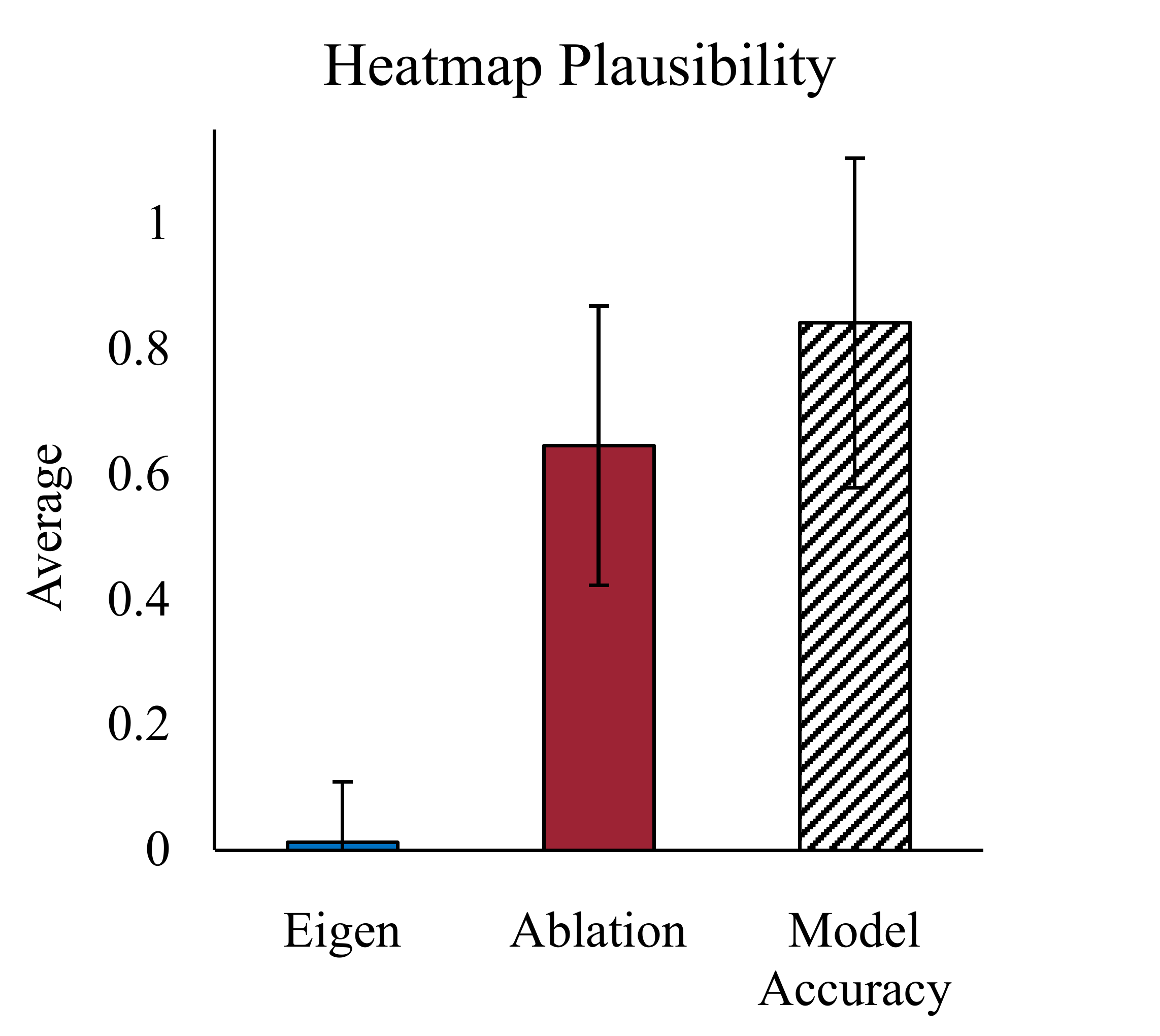}} \\
            & \multicolumn{1}{c}{High Plausibility} \\
            & \includegraphics[width=0.32\linewidth]{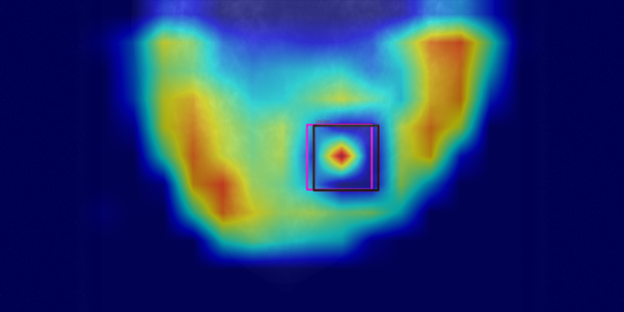} \\
            & \multicolumn{1}{c}{Low Plausibility} \\
            & \includegraphics[width=0.32\linewidth]{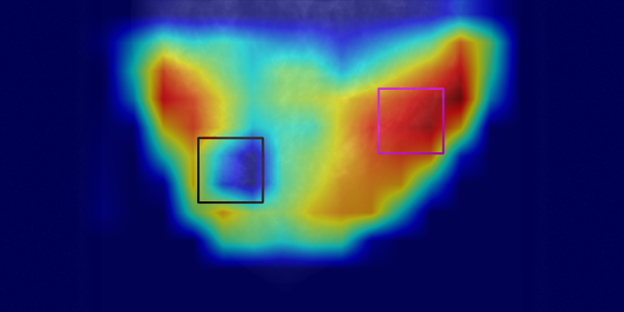} 
    \end{tabular}
    
    \caption{A) Results for Plausibility. Metrics are calculated using Intersection Over Union (IOU) and Model Accuracy. B) Examples of high and low plausibility heatmaps. Pink squares are the ROI centered in the highest activation area. Black squares are the ground truth lesion location ROI.}
    \label{fig:results-main2}
\end{figure}

Figure~\ref{fig:results-main2} illustrates examples of heatmaps with varying levels of plausibility. Figure~\ref{fig:results-main2}.A presents the quantitative plausibility results, where IoU and Model Accuracy were used to calculate the average plausibility scores. As shown in the plot, Ablation CAM demonstrates a significantly higher plausibility score compared to Eigen CAM, indicating a better alignment with ground truth lesion locations. The lower plausibility of Eigen CAM can be attributed to how this method computes the principal components of feature maps, which provides a more global representation of the image’s features. While this approach captures broader activation patterns, it may not always localize the specific regions most relevant to the clinical task, such as detecting small lesions. In contrast, Ablation CAM systematically removes parts of the model to assess each component's contribution to the prediction, allowing it to focus more on localized regions that directly influence the model’s decision.

Figure~\ref{fig:results-main2}.B provides visual examples of high and low plausibility heatmaps. In the high plausibility example, the heatmap aligns closely with the ground truth lesion location (represented by the black square), and the region of interest (ROI) in the heatmap (indicated by the pink square) is centered on the lesion, demonstrating good alignment between the model’s explanation and human expectations. Conversely, the low plausibility example shows poor alignment between the heatmap's highest activation area (pink square) and the true lesion location (black square), with the highlighted region failing to correspond accurately to the lesion. These results show that some explainability methods can be less plausible, even if they are consistent. Therefore, after assessing the consistency of explainability methods, it is important to ensure they provide explanations that align with human ground truth. 

A comprehensive evaluation of plausibility may require incorporating different levels of ground truth. For example, in mammography, the most direct ground truth might be the lesion region itself, as annotated by expert radiologists. At the same time, indirect ground truth, such as surrounding anatomical structures or contextual imaging features, also constitutes valuable information, as humans often consider these regions when making decisions. This information can be obtained through tools like eye-tracking, which capture whether radiologists focus not only on lesion regions but also on adjacent areas that provide contextual clues about a lesion’s characteristics. By incorporating this hierarchy of ground truth—from precise lesion annotations to indirect contextual cues—a more nuanced and comprehensive evaluation of plausibility can be achieved.

\subsubsection{Fidelity}
Fidelity measures the agreement of an explanation with the underlying mechanisms the model used to predict an outcome. This concept is also referred to as faithfulness or truthfulness. A high-fidelity explanation accurately reflects the factors the model considered important in its decision. For example, in an image-based task, the pixels representing the detected object should be highlighted if they are informative enough for the model to make its prediction, while non-relevant pixels should remain unhighlighted. It is important to note that model's accuracy and fidelity are related but distinct concepts. Fidelity refers to how well the explanation aligns with the internal workings of the model, while accuracy measures how well the model performs at its task. If a model is highly accurate, and the explanation has high fidelity, this means the explanation accurately reflects the true reasons behind the model's predictions. However, some explanations may only provide local fidelity, meaning they accurately represent the model’s reasoning for a specific subset of the data (e.g., a particular instance or a small group of examples). For example, local surrogate models or Shapley Values explain individual predictions by focusing on how the model behaves for that particular instance, rather than providing a global understanding of the entire model's decision-making process.

Several approaches exist to measure the fidelity of an explainability method. One well-known approach, introduced by \citeauthor{adebayo2018sanity} \cite{adebayo2018sanity}, is the Model Parameter Randomization Check. In this method, the model's parameters are perturbed by randomizing or re-initializing the weights, which disrupts the original learned patterns. Explanations are then generated both before and after this randomization process. The key idea is to compare the explanations: if the explanations remain the same after randomization, it suggests that the explanation is not sensitive to the model's learned parameters and does not faithfully reflect the model’s reasoning. On the other hand, if the explanations differ after randomization, it indicates that the explanation is indeed tied to the model's internal mechanisms. Another method for assessing fidelity is the White Box Check \cite{nauta2023anecdotal}, which evaluates explainability methods by using a transparent model with known decision-making processes, such as a decision tree or random forest. The generated explanations are compared to the model's internal reasoning, allowing for a direct assessment of how accurately the explanation method reflects the model's true decision-making process.

Additional methods include the Single Deletion Method and the Incremental Deletion Method \cite{nauta2023anecdotal}. The Single Deletion Method evaluates feature importance by testing how much the model's output changes when a single feature is removed or altered. A correct explanation should show that removing the most important feature leads to the largest change in the model's output, while removing less important features causes minimal impact. The Incremental Deletion Method extends this by removing features one by one, either in order of importance or in reverse, to assess how the model's output changes. This approach helps verify whether the cumulative removal of features corresponds to their ranked importance.

\begin{figure}[t]
    \centering
    \begin{tabular}{ll}
    {\large A) Fidelity} & \\
         \multicolumn{2}{c}{\includegraphics[trim={0cm 0cm 0cm 3.5cm},clip,width=0.55\linewidth]{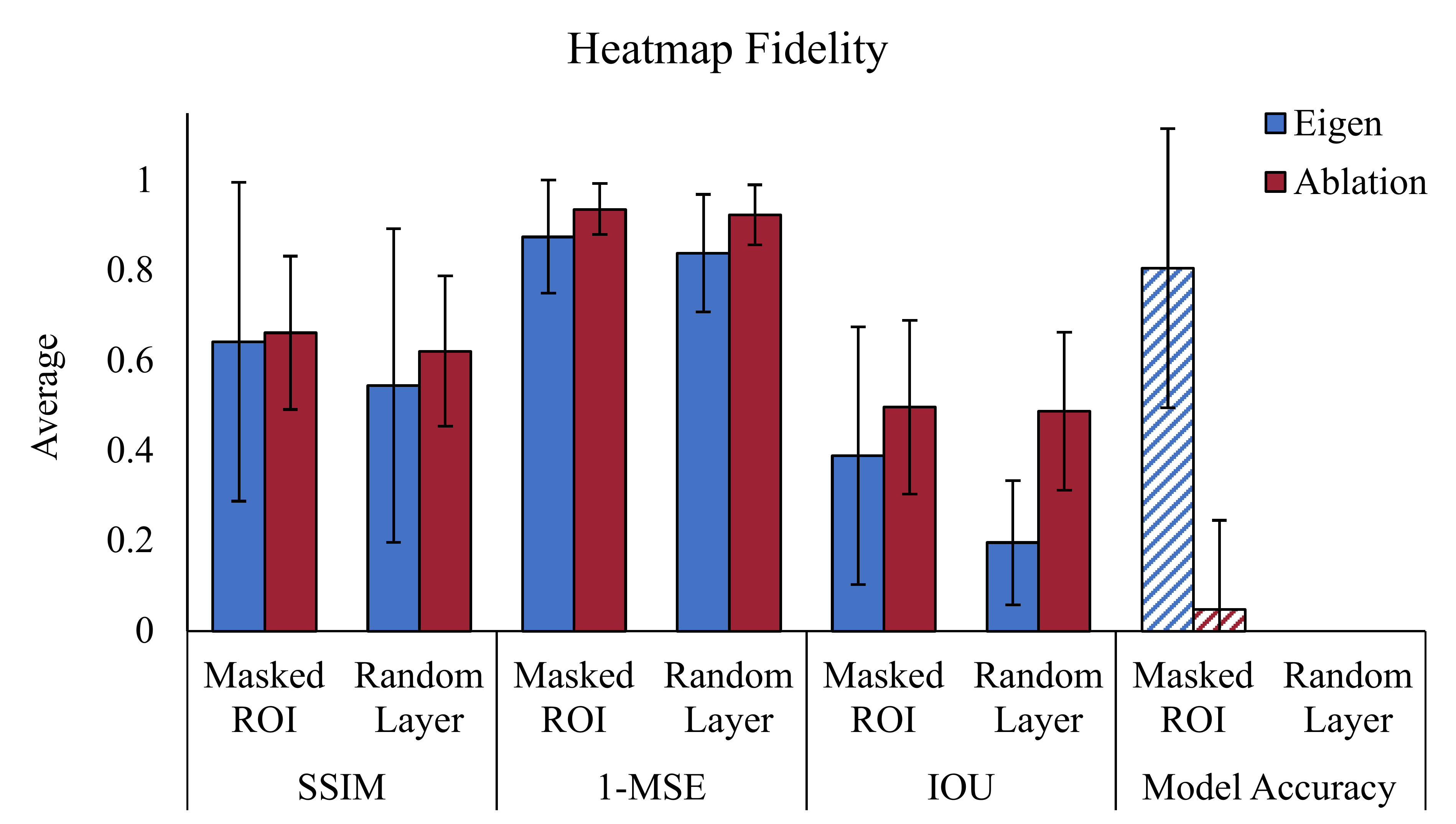}}  \\ 
         {\large B) Examples}&\\
         \multicolumn{1}{c}{No change in model prediction} & \multicolumn{1}{c}{No change in model prediction}\\
         \multicolumn{1}{c}{Small change in heatmap} & \multicolumn{1}{c}{Large change in heatmap}\\
         \includegraphics[width=0.45\linewidth]{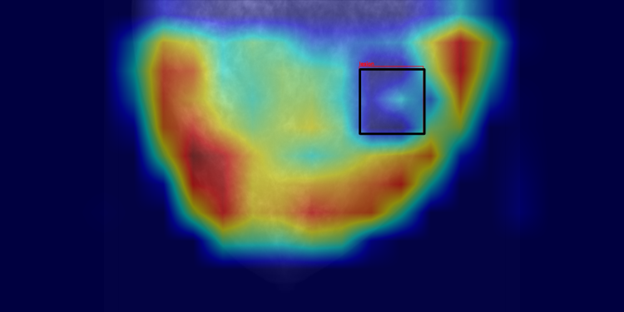} & \includegraphics[width=0.45\linewidth]{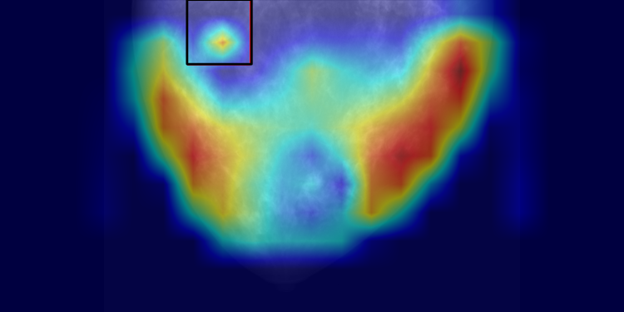} \\
         \includegraphics[width=0.45\linewidth]{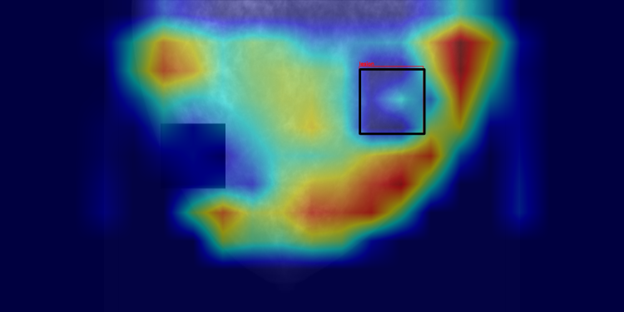} &
         \includegraphics[width=0.45\linewidth]{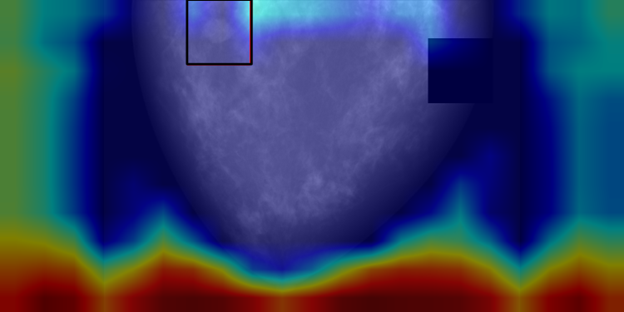}
    \end{tabular}
    
    \caption{A) Results for Fidelity. Metrics are calculated using  Structural Similarity Image Metric (SSIM), Mean Square Error (MSE), Intersection Over Union (IOU), and Model Accuracy. B) Examples of low (left) and high (right) fidelity heatmaps.}
    \label{fig:results-main4}
\end{figure}

Figure~\ref{fig:results-main4} illustrates the fidelity of two saliency map methods. As shown in Figure~\ref{fig:results-main4}.A, Ablation CAM consistently exhibits higher fidelity than Eigen CAM, indicating that it more faithfully reflects the model’s internal reasoning when the model's parameters are perturbed. Particularly, with each metric, Ablation CAM has reduced similarity to the un-perturbed explanation, but correspondingly reduces the accuracy of the model, unlike Eigen CAM. Together, this suggests that Ablation CAM is producing high fidelity explanations due to disruption in similarity and model performance. In contrast, Eigen CAM shows lower fidelity, with increased variability between the different metrics and a spared model accuracy after perturbation for the masked ROI case, which highlights its weaker connection to the model's internal mechanisms. Together, these results suggest that comparison between explanations before and after perturbation is necessary to examine fidelity but that examining accuracy is also critical for a full account of fidelity. Though fidelity and model accuracy are independent, understanding how the explanation aligns with the model's decision is critical to fully understanding fidelity. The pattern of results presented in Figure~\ref{fig:results-main4} is a great example of this need for both pieces of information, as both explanations were disrupted by perturbations, but only Ablation CAM resulted in a more consistent disruption and a change in the model's decision making. However, if just examining model accuracy, in the case of randomizing model layers, we would suspect that both explanations have high fidelity, which is not the case. 

Figure~\ref{fig:results-main4}.B offers visual examples that further illustrate these findings when we removed the highest activation ROI from the original input (top) and fed it back through the model (bottom). On the left column, the example shows no change in the model prediction and only a small change in the heatmap, indicating that the explanation is not affected even after minor perturbations to the input, which reflects lower fidelity. On the right column, another example shows no change in the model prediction but a large change in the heatmap, suggesting that the explanation is sensitive to perturbations based on the highest heatmap activation areas, indicating higher fidelity.

It is important to note the difference between plausibility and fidelity. Plausibility focuses on measuring the alignment between the explanation and domain knowledge or ground truth, while fidelity focuses on measuring the alignment of the explanation with respect to the predictive model. For example, an explanation highlighting patient information in the background to distinguish between sick and healthy patients would score low on plausibility with respect to tumor segmentation masks but could still correctly reflect the reasoning of this flawed classifier (high fidelity). Therefore, it is important to evaluate explainability methods for both plausibility and fidelity independently.

\subsubsection{Usefulness}

The final evaluation component focuses on assessing the explanation's ability to help the user in effectively performing the given task. This involves two aspects: Clinical Relevance and Interpretability.

Clinical relevance refers to the degree to which an explanation is useful and informative for the specific medical task it is intended to support. It goes beyond whether the explanation merely aligns with general ground truth (as in plausibility) and instead focuses on whether the explanation directly addresses the critical aspects of the clinical task. This criterion evaluates if the information provided by the model is meaningful and necessary for clinicians to make informed decisions. For example, in a breast density evaluation task, clinical relevance would require that the model's explanation focuses on the dense tissue areas in a mammogram, as these regions are essential for assessing breast density. An explanation that highlights irrelevant or peripheral areas may still seem plausible in some contexts, but would lack clinical relevance because it does not inform the task at hand. Thus, clinical relevance ensures that the explanation contributes directly to the decision-making process and enhances the clinician’s ability to interpret the model’s output in a medically useful way.

Interpretability refers to the extent to which a non-AI expert (e.g., a radiologist) can extract and understand information from the provided explanation. It is often synonymous with understandability and concerns how easily the internal workings of the model can be made accessible and understandable to end users. The National Institute of Standards and Technology (NIST) attempts to draw a distinction between interpretation--the process of deriving meaning from a stimulus--and explanation--the process of generating detailed description of how the outcome was reached--\cite{broniatowski2021psychological-NIST}, and we generally agree with this distinction. Interpretability focuses on how well the model is designed to be understood by human users. A model can still be difficult to interpret if the information processing is too complex for the user to comprehend (e.g., deep neural network). 

The concepts of clinical relevance and interpretability are related, as both focus on how well an AI-generated explanation can assist human users, especially non-AI experts, in understanding and making decisions based on the model’s output. Clear and easy-to-understand explanations (high interpretability) are more likely to be clinically relevant, as they reduce cognitive load and enhance decision-making. However, an explanation can be interpretable but not clinically relevant if it highlights the wrong areas or fails to assist with the specific task. Similarly, a clinically relevant explanation can be hard to interpret if the model's inner workings are too complex. Low interpretability can negatively impact a user's performance, thus reducing trust in the system. Superficial or unrelated explanations can increase a user's cognitive load and decrease the efficiency and/or performance of the user. 



Assessing usefulness is only possible through studies in which medical professionals evaluate the generated explanations for their usefulness. The human evaluation should focus on interpretability \cite{10.5555/3504035.3504222} or time-to-decision improvement \cite{lakkaraju2016interpretable}. One possibility is that users are asked to perform the selected task both with and without the provided explanation; if the provided explanation is useful, it should either enable enhanced performance on the task or completion of the task more quickly. 

\subsection{From Quantitative to Human-Based Assessment}
We have defined four criteria for evaluating explainability methods: Consistency, Plausibility, Fidelity, and Usefulness. The order of evaluation is important. Consistency must be evaluated first, as it ensures that the explainability method provides stable and reproducible explanations across similar inputs and conditions. Once consistency is confirmed, plausibility is assessed to measure how well the explanation aligns with expert annotations. By evaluating plausibility early, we ensure that the explanation aligns with human perspective. If the explanation fails this test, it may lack meaningful insight, regardless of its connection to the model’s internal mechanisms. After plausibility, fidelity is assessed to determine whether the explanation accurately reflects the model's reasoning. Even if an explanation is plausible to human experts, it can still be misleading if it does not correctly represent how the model arrived at its decision. 

Once an explainability method passes Consistency, Plausibility, and Fidelity, it is ready to move into the second stage for human-based evaluation, where medical experts assess the explanation's usefulness for specific clinical tasks. Requiring a method to pass consistency, plausibility, and fidelity before involving human experts, filters out poor or misleading explanations, ensuring that only a subset of explanations require human evaluation. 

\section{Discussing Explainability}
While explainability features are desired by users of AI-enabled devices, there is currently a lack of concrete definition of explainability as well as evaluation techniques for assessing the quality of provided explanations. This lack of concrete definitions and techniques for assessing explanations is problematic, as a growing body of evidence suggests that providing untrustworthy information to users can degrade task performance \cite{fuxin2021heatmaps, bernstein2023can, jabbour2023measuring,khera2023automation}. In this work, we propose a comprehensive framework for the evaluation and reporting of explainable AI features. We demonstrate the feasibility of our proposed framework by applying it to heatmap-based explanations for medical images. However, our intention is for this framework to be broadly applicable to different explainable AI methods. We hope our work can begin the dialogue of how best to evaluate explanations provided by AI. The scorecard included in the supplementary material serves as an example of how to report the information for one specific explanation method.

Three of our proposed criteria--consistency, plausibility, and fidelity--can be evaluated retrospectively on previously-collected data. On the contrary, Usefulness require a human task-based performance assessment. To best use resources, we have structured the evaluation framework in a specific order, proposing the evaluation of Consistency, Plausibility, and Fidelity first, followed by Usefulness. This ordering helps to minimize resources, as explanations with low consistency, plausibility and fidelity need not proceed to human task-based evaluation studies.  

To the best of our knowledge, no clear consensus exists on what makes an explanation `good' or `useful.' The ambiguity of this question is at least partially explained by the fact that the interpretation of an explanation is linked to the context in which that explanation is provided. In the current work, we have framed a good explanation as one that consistently provides the same information, is aligned with the ground truth, changes with the output of the model, and is useful for the clinical task. This framing is consistent with prior thinking that an ideal explanation must convey the \textit{gist} of what is needed, but without overwhelming with irrelevant information \cite{broniatowski2021psychological-NIST}. Our work extends prior work that explanations should display fidelity, interpretability, and be independent of any specific AI model \cite{Celino_2020}. 

A definite confounder is that different users will have different needs, abilities and desires from explanations \cite{Celino_2020}. Put another way, explanations are useful to different groups for different reasons, and one can further argue that explanations are most useful when they are tailored to their desired intended user \cite{COMBI_2022}. AI developers generally seek explanations to debug or improve their models. For the developer, a good explanation is causal and helps illustrate why something is occurring (e.g. the algorithm shows bias towards a particular group because the training data was imbalanced), so that additional action can be taken (e.g. collecting more training data). A user interacting with the same model would be more interested in whether or not the model output could be relied upon as fair, and their potential actions would be to use or not use the model output.

AI-provided explanations must be interpretable by human users.  A significant hurdle to achieving this is that humans and computers process information differently. Whereas humans rely on simple, imprecise gists to make decisions, machine learning models rely on programmatic verbatim processes to generate predictions \cite{broniatowski2021psychological-NIST}. Translating the precise technical decisions of a computer into human-interpretable explanations is an ongoing challenge.  Additionally, the vast majority of AI-models deployed in healthcare today are used as assistive devices, meaning that the impact of the model is reliant on the combination of the AI model output and the human interpretation. This adds an additional layer of complexity, as users differ in the degree to which they are willing and able to utilize their own background knowledge to interpret detailed technical information, and users place different amounts of trust in AI output.

It is noteworthy to mention that this framework does not provide a threshold or any acceptability criteria. However, the framework offers enough flexibility to include acceptance thresholds for each of the four criteria. These thresholds must be defined and justified independently for different applications. Some AI algorithms, depending on the task, may prioritize high consistency whereas others may focus on better fidelity. 

 
\section{Conclusion}
Explanations provided by AI systems are seen as a means of increasing trust and confidence in the AI systems. However, to be useful, such explanations must provide only useful information that is interpretable to the user. Unfortunately, a comprehensive evaluation framework for AI explanations is not currently available. To fill this gap, we propose a framework for the evaluation of explainability features in AI-models applied to medical data. Our framework is based on four criteria that evaluate 1) consistency of explanations, 2) plausibility of explanations, 3) fidelity of explanations, as well as 4) usefulness. We also propose an explainability scorecard (included as supplementary material) for standardized reporting of the results. 




\bibliography{report}   
\bibliographystyle{plainnat}   



\end{spacing}
\end{document}